%% file: ms.tex
\documentclass[11pt,a4paper]{article}
\usepackage[hyperref]{swisstextkonvens2020}
\usepackage{times}
\usepackage{latexsym}
\usepackage{color,soul}
\usepackage{enumitem}
\usepackage{verbatim}

\usepackage{listings}
\makeatletter
\newcommand{\mylabel}[2]{#2\def\@currentlabel{#2}\label{#1}}
\makeatother

\hypersetup{breaklinks}
\usepackage[hyphenbreaks]{breakurl}

\usepackage{arydshln} 

\usepackage{amssymb}
\usepackage{multirow}

\usepackage{etoolbox}
\apptocmd{\thebibliography}{\raggedright}{}{}

\usepackage{array,graphicx}
\usepackage{ragged2e}
\usepackage{float}
\usepackage{xcolor}
\usepackage{microtype}

\usepackage{booktabs}

\aclfinalcopy 

\title{Evaluating German Transformer Language Models \\ with Syntactic Agreement Tests}
\usepackage{amsmath}
\newcommand\blfootnote[1]{%
  \begingroup
  \renewcommand\thefootnote{}\footnote{#1}%
  \addtocounter{footnote}{-1}%
  \endgroup
}

\author{Karolina Zaczynska*, Nils Feldhus*, Robert Schwarzenberg, \\ \textbf{Aleksandra Gabryszak}, \textbf{Sebastian M\"oller} \\
  German Research Center for Artificial Intelligence (DFKI) \\
  \texttt{\{firstname.lastname\}@dfki.de}
  }

\date{}

\usepackage{amsthm}
\theoremstyle{definition}
\newtheorem{exmp}{Example}[section]

\newtheoremstyle{StyleForTextExamples}
{}
{}
{\ttfamily\small}
{}
{\itshape\small}
{}
{\parskip}
{}
\theoremstyle{StyleForTextExamples}

\begin{document}

\maketitle

\input{sections/abstract}
\input{sections/1_introduction}

\input{sections/2_methods}
\input{sections/3_dataset}
\input{sections/4_experiments}
\input{sections/5_resultsanddiscussion}
\input{sections/6_related_work}
\input{sections/7_conclusion}
\input{sections/acknowledgements}
\input{sections/references}

\end{document}

%% file: sections/abstract.tex
\begin{abstract}

Pre-trained transformer language models (TLMs) have recently refashioned natural language processing (NLP): Most state-of-the-art NLP models now operate on top of TLMs to benefit from contextualization and knowledge induction. To explain their success, the scientific community conducted numerous analyses. Besides other methods, syntactic agreement tests were utilized to analyse TLMs. Most of the studies were conducted for the English language, however. In this work, we analyse German TLMs. To this end, we design numerous agreement tasks, some of which consider peculiarities of the German language. Our experimental results show that state-of-the-art German TLMs generally perform well on agreement tasks, but we also identify and discuss syntactic structures that push them to their limits.

\blfootnote{* Shared first authorship.}
\end{abstract}

%% file: sections/1_introduction.tex
\section{Introduction}

Pre-trained language models, in particular those which are based on the transformer architecture \cite{vaswani2017attention}, have immensely improved the performance of various downstream models (see, e.g. \citet{zhang2020retrospective, zhang2019semantics, raffel2019exploring}). To explain their success, numerous introspective experiments have targeted different aspects of TLMs. It was shown, for instance, that they encode syntactic, semantic and world knowledge \cite{petroni_language_2019} and present downstream models with a highly contextualized representation of the input tokens \cite{tenney_what_2019}. For a comprehensive overview of the many studies conducted about arguably the most prominent of language models, \textsc{Bert} \cite{devlin2019bert}, we refer the interested reader to the excellent overview paper by \citet{rogers2020primer}.

With the exception of experiments targeting a multilingual \textsc{Bert} model \cite{rogers2020primer}, most of the studies were conducted only for English, however. Other languages are underrepresented. In this work, we narrow the gap for German by analysing the abilities and limits of German TLMs. To the best of our knowledge, we are the first to conduct such an analysis for the German language.

When compared with English, there are considerable syntactic differences in the German language that we consider in this work. For example, the inflection system of the German language is more complex, the range of morpho-syntactic rules needed to form grammatical sentences is larger, and the allowed word order is more diverse. As a consequence, the German language models face specific challenges. The syntactic agreement tests presented in this work include several of them.

Our main contributions are threefold:

\begin{enumerate} \RaggedRight
    \item Utilizing context-free grammars (CFG), we compile a German data set of controlled syntactic correctness tests of various complexities. The motivation and construction of the data set is closely following the one described in \citet{marvin-linzen-2018}, where syntactic tests were conducted for English. In particular, we devise several kinds of subject-verb agreement as well as reflexive anaphora agreement tasks, taking into account peculiarities of the German language. A simple subject-verb agreement task is given in Example~\ref{exmp:simpleagreement}.
    
    \begin{exmp}
    \label{exmp:simpleagreement}
    Decide which of the following sentences is grammatical:
    \begin{enumerate}[topsep=0px,noitemsep,label=(\alph*)]
        \item Der Autor \textit{lacht}. (The author laughs.)
        \item * Der Autor \textit{lachen}. (The author laugh.)
    \end{enumerate}
    \end{exmp}

    \item We use the data set to evaluate two transformer-based language models that were pre-trained on German corpora. During the evaluation, contrary to prior work, we utilize the cross entropy loss to score the syntactic correctness of input sentences. This addresses a problem with the sub-word tokenization of some TLMs that was previously solved by discarding thousands of data points.

    \item We conduct a qualitative and quantitative analysis of the experimental results, 
    estimating the abilities and limits of the TLMs tested. 
\end{enumerate}

%% file: sections/2_methods.tex
\section{Methods}
\label{sec:methods}
Our work combines and translates the targeted syntactic evaluation of language models by \citet{marvin-linzen-2018} and the assessment of \textsc{Bert}'s syntactic abilities by \citet{goldberg_assessing_2019} from English into German. Our methods consist of agreement test generation and model evaluation. 

We created the following agreement test, following \citet{marvin-linzen-2018}: Two sentences, a grammatical one and an ungrammatical one, are forwarded through a model. The sentences differ minimally from each other at only one locus of (un)grammaticality, i.e. one word. The model output is monitored and if the output suggests that the model prefers the grammatical one over the ungrammatical one, that instance is counted as a correct classification; otherwise, it is counted as an incorrect classification. 

\citet{goldberg_assessing_2019} used agreement tests to evaluate \textsc{Bert} models. To account for their bidirectionality, he masked the locus of (un)grammaticality and queried the candidate probabilities for the mask.  In Example~\ref{exmp:simpleagreement}, \texttt{Der Mann [MASK].} is forwarded through a \textsc{Bert} model and the candidate probabilities at the position of the mask are determined. If \texttt{lacht} receives a higher probability than \texttt{lachen}, the task is solved correctly by the language model.
The author runs into problems, however, when the candidates are tokenized into multiple sub-word tokens, say \texttt{lachen} $\rightarrow$ [\texttt{lach, \#\#en}]. In this case, the author simply ignores the data point. 

Instead of discarding such sequences, we take inspiration from \citet{marvin-linzen-2018} and score whole sentences (without masks). However, we still discard cases in which the two candidates have a different amount of sub-words after tokenization, as we see the comparability impaired if the resulting sequences of tokens are of different lengths.

We compute the sentence score with the cross-entropy loss of the forward pass, using the input sequence as the target:

\begin{equation*}
    \label{eq:loss}
    \frac{1}{T}\sum_{i=1}^{T}\left(-f(S)_{i,S_{i}}+\log (\sum_{j=1}^{T}\exp(f(S)_{j,S_{j}}))\right)
\end{equation*}
where $S$ is a sequence of $T$ positive integer token ids and $f: \mathbb{Z}^{N} \rightarrow \mathbb{R}^{N \times V}$ a language model mapping $N$ token IDs onto $N$ token probabilities over a vocabulary of size $V$. We compute Eq.~\ref{eq:loss} with the grammatical candidate in place and a second time with the ungrammatical candidate in place.

Please note that during the training of a bidirectional language model, the points of interests need to be masked to prevent information leakage \cite{devlin2019bert}. In our case, information leakage is not a problem because we compare two whole sequences.

%% file: sections/3_dataset.tex
\section{Syntactic Agreement Tests}

This section describes the syntactic agreement tests we generated to evaluate German TLMs on.

Our tests are inspired by the research of \citet{marvin-linzen-2018} and \citet{goldberg_assessing_2019}. In particular, we translate many of their tests on subject-verb agreement (SVA) and reflexive anaphora (RA) agreement from English to German (Section~\ref{sec:established_tests}). In addition, we design tests for syntactic phenomena which are typical of the German language (Section~\ref{sec:novel_tests}). 

The generated tasks cover a range of difficulties. In  German, the subject and the inflected verb agree with regard to person and grammatical number. In the simplest case, the sentences contain only a subject and a verb. In the more challenging cases we added different types of distraction, i.e. either additional non-subjective (pro)nouns as candidates for subjects or other additional lexical material making the sentences more complex. 

For our experiments, we consider instances where the grammatical number of non-subjective (pro)nouns matches the one of the subject as well as examples where their grammatical number is different.
Furthermore, we distinguish between local and non-local feature agreement, which means, we take into account whether the distractors occur between subject and its corresponding verb or not. The described test scenario allows us to compare the models' performance with regard to the features of the distractor as well as its distance to the relevant verb.  Therefore, the designed tests expand the experimental setup of \citet{marvin-linzen-2018} by going beyond the attractors, i.e. intermissions defined as intervening nouns with the opposite number from the subject \cite{linzen-etal-2016-assessing}.

\subsection{Dataset}
\label{sec:dataset}

We created a dataset of 13,002 sentences using hand-crafted Context Free Grammars (CFGs) as illustrated in Example~\ref{exmp:cfg}.

\begin{exmp}
\label{exmp:cfg}
\textit{Context Free Grammar} for creating sentences \emph{S} from a vocabulary \emph{V} to test agreement in a simple sentence:
\begin{lstlisting}[columns=fullflexible]
    S -> NP V '.'
    NP -> ART N
    ART -> 'Die'
    N -> 'Autoren' | 'Richterinnen'
    V -> 'lachen' | 'reden'
\end{lstlisting}
\textbf{Output:} \textit{Die Autoren lachen. / Die Autoren reden. / Die Richterinnen lachen. / Die Richterinnen reden.}

\end{exmp}

As shown in the example, the CFG creates sentences as output with varying lexical items but with a relatively low variance. However, it allows us to tightly control the generated sentences with respect to the desired tests, in terms of distractor features as well as syntactic structure and correctness of the sentences.

Our data set covers 14 test cases of different challenge levels (Sections~\ref{sec:established_tests}--\ref{sec:novel_tests}). The number of sentences ranges from 64 to 2160 with an average of 928,71 sentences per test case. A sentence is build on average of 6.88 tokens. The vocabulary consists of 88 lexems and 171 word forms. For our corpus, we chose common words to build the sentences, so that the TLM was not confronted with potentially unknown words.

\subsection{Established Agreement Tests}
\label{sec:established_tests}

In the following, we introduce the agreement tests that we translated from the work of  \citet{marvin-linzen-2018}. 

We describe three groups of tests ordered by the increasing challenge level: (1) local agreement, no distractors, (2) local agreement, plus distractors, and (3) non-local agreement, plus distractors. Afterwards, we introduce tests designed to target German phenomena specifically. 

\paragraph{Local agreement, no distractors}
 We first include cases with local agreement and without a distractor. Sentences consisting of only one subject and verb are what we refer to as \textit{simple sentence} in the following, showcased in Example~\ref{exmp:localagreement}.

\begin{exmp}
\label{exmp:localagreement}
\textit{Simple sentence} with only one subject and one verb (the locus of (un)grammaticality is italic, the incorrect variant is preceded by *):
\begin{enumerate}[topsep=0px,noitemsep,label=(\alph*)]
    \item Das Kind \textit{trinkt}.
    \item * Das Kind \textit{trinken}.
\end{enumerate}
\end{exmp}

\paragraph{Local agreement, plus distractors}

Complex sentences with a local agreement in a \textit{sentential complement} or in an \textit{object relative clause} constitute the next level of difficulty. Those sentences contain two subjects: one in the main clause, and another one in the subordinate clause. In Example~\ref{exmp:sentinential_complement}, the latter functions as a sentential complement, in Example~\ref{exmp:inobjrelative}, as an object relative clause. For both types of the subordinate clause, the verb follows the subject directly. The subject of a main clause is the distractor in those cases while the agreement between the subject and the verb of the subordinate clause is our point of interest.   

\begin{exmp}
\label{exmp:sentinential_complement}
\textit{SVA in a sentential complement}:

\begin{enumerate}[topsep=0px,noitemsep,label=(\alph*)]
    \item  Die Vertreter sagten, dass das Kind  \textit{trinkt}.
    \item * Die Vertreter sagten, dass das Kind \textit{trinken}.
\end{enumerate}
\end{exmp}

\begin{exmp}
\label{exmp:inobjrelative}
\textit{SVA in an object relative clause}

\begin{enumerate}[topsep=0px,noitemsep,label=(\alph*)] 
    \item Der Autor, den die Vertreter \textit{kennen}, lacht.
    \item * Der Autor, den die Vertreter \textit{kennt}, lacht.
\end{enumerate}
\end{exmp}
\noindent

\paragraph{Non-local agreement, plus distractors}

We also tested TLMs on a set of constructions with non-local agreement, induced by potentially distracting words and phrases between the head of the subject and its corresponding verb. With these tasks, we are testing the language model's ability to attend to the subject in sentences across long contexts. 

Our first test case is a \textit{SVA across a preprositional phrase} (PP). We created sentences with the subject modified by a directly following PP, which includes a potentially attracting noun, as in Example~ \ref{exmp:svaacrossprep}. 

\begin{exmp}
\label{exmp:svaacrossprep}
\textit{SVA across a PP}

\begin{enumerate}[topsep=0px,noitemsep,label=(\alph*)]
    \item  Der Autor neben den Landstrichen  \textit{lacht}.
    \item * Der Autor neben den Landstrichen  \textit{lachen}.
\end{enumerate}
\end{exmp}

Furthermore, we test \textit{SVAs across subject relative clauses} which include one potentially distracting object and verb in between subject and corresponding verb, as in Example~\ref{exmp:subrelclause}.

\begin{exmp}
\label{exmp:subrelclause}
\textit{SVA across a subject relative clause}

\begin{enumerate}[topsep=0px,noitemsep,label=(\alph*)]
    \item Der Autor, der die Architekten liebt,  \textit{lacht}.
    \item * Der Autor, der die Architekten liebt,  \textit{lachen}.
\end{enumerate}
\end{exmp}

The same challenge exists for \textit{SVAs across object relative clauses} which also contain potentially distracting chunks and separate the subject and its corresponding verb, as in Example~\ref{exmp:nonlocalagreement}.

\begin{exmp}
\label{exmp:nonlocalagreement}
\textit{SVA across an object relative clause}

\begin{enumerate}[topsep=0px,noitemsep,label=(\alph*)] 
    \item Der Autor, den die Vertreter kennen, \textit{lacht}.
    \item * Der Autor, den die Vertreter kennen, \textit{lachen}.
\end{enumerate}
\end{exmp}
\noindent

Additionally, we designed various sentences for testing \textit{SVAs across coordinated verbal phrases} (VP), where the subject must agree in person and number with the finite verb included in each VP. In our test, the point of interest is the second verb of the coordination. This kind of structure challenges the model to recognize that the complete subject-verb structure does not end after the first verb, but rather it also includes the second verb. We test the SVA in verbal coordinations of different lengths and various number of distractors.

First, we test the model on sentences consisting of a short and simple VP coordination with no distractors, as illustrated by Example~\ref{exmp:verbalcoord_simple}.

\begin{exmp}
\label{exmp:verbalcoord_simple}
\textit{SVA in short VP coordinations (i.e. with no distractors)}

\begin{enumerate}[topsep=0px,noitemsep,label=(\alph*)]
    \item Der Autor schwimmt und \textit{lacht}.
    \item * Der Autor schwimmt und \textit{lachen}.
\end{enumerate}
        
\end{exmp}

To increase the difficulty level, we inserted noun phrases having a different number as the subject into the coordinated VP. We distinguish between verbal coordinations with a single noun distractor (Example~\ref{exmp:verbalcoord_short_distractor1}) and two noun distractors (Example~\ref{exmp:verbalcoord_long_distractor2}). 

\begin{exmp}
\label{exmp:verbalcoord_short_distractor1}
\textit{SVA in medium VP coordinations (i.e. with a single noun distractor)}

\begin{enumerate}[topsep=0px,noitemsep,label=(\alph*)]
    \item  Der Autor redet mit Menschen und \textit{lacht}.
    \item * Der Autor redet mit Menschen und \textit{lachen}.
\end{enumerate}
\end{exmp}

\begin{exmp}
\label{exmp:verbalcoord_long_distractor2}
\textit{SVA in long VP coordinations (i.e. with two noun distractors)}

\begin{enumerate}[topsep=0px,noitemsep,label=(\alph*)]
    \item Der Autor redet mit Menschen und \textit{verfolgt} die Fernsehprogramme. 
    \item Der Autor redet mit Menschen und \textit{verfolgen} die Fernsehprogramme.
\end{enumerate}
\end{exmp}

\subsection{Novel Agreement Tests}
\label{sec:novel_tests}

In addition to the tests above that we based on previous work, we also designed tasks which target constructs that are more specific to the German language. 
\label{ssec:germanscases}

First, we test the agreement between verb and its corresponding subject containing an extended modifier, i.e. an adjective modifying a subject and extended by further subordinate nominal or prepositional phrase. The extended modifier is positioned between the determinator and the noun of the subject. In comparison to English, the German language is much more flexible with regard to the number and the types of allowed extensions. To test the impact of nouns used within extended modifiers of a subject we include sentences with simple modifiers as well as with extended modifiers (Example~\ref{exmp:short_verbal_modifier} and \ref{exmp:verbal_modifier}). 

\begin{exmp}
\label{exmp:short_verbal_modifier}

\textit{SVA with a simple modifier}

\begin{enumerate}[topsep=0px,noitemsep,label=(\alph*)]
    \item Die wartenden Autoren \textit{lachen}. 
    \item * Die wartenden Autoren \textit{lacht}.
\end{enumerate}
\end{exmp}

\begin{exmp}
\label{exmp:verbal_modifier}

\textit{SVA with an extended modifier}

\begin{enumerate}[topsep=0px,noitemsep,label=(\alph*)]
    \item Die die Pflanze liebenden Autoren \textit{lachen}. 
    \item * Die die Pflanze liebenden Autoren \textit{lacht}.
\end{enumerate}
\end{exmp}

\begin{table*}[htb!]
    \centering
    \begin{tabular}{lccc}
        \toprule
        & $_{\text{distil}}$\textsc{gBert} & \textsc{gBert}$_{\text{large}}$ & \# sents \\
        
        \textsc{Subject-verb agreement} & & & \\
        
        Simple Sentence
        & \textbf{0.9710} & 0.9420 & 69 \\

        In a sentential complement
        & 0.9565 & \textbf{0.9894} & 2160 \\

        Short VP coordination
        & \underline{0.7125} & \underline{\textbf{0.7542}} & 240 \\

        Medium VP coordination
        & \underline{0.4813} & \underline{\textbf{0.6188}} & 480 \\

        Long VP coordination
        & \underline{0.5167} & \underline{\textbf{0.5938}} & 480 \\

        Across a PP
        & 0.7968 & \textbf{0.9005} & 2160 \\

        Across a subject relative clause
        & \underline{0.6924} & \textbf{0.9896} & 1440 \\

        Across an object relative clause
        & 0.7386 & \textbf{0.9206} & 945 \\
 
        In an object relative clause
        & 0.9568 & \textbf{0.9600} & 1575 \\
        
        With a modifier
        & 0.9458 & \textbf{0.9959} & 240 \\
        
        With an extended modifier
        & 0.8917 & \textbf{0.9583} & 480 \\

        Pre-field
        & 0.73 & \underline{\textbf{0.7987}} & 348 \\

        \midrule
        
        \textsc{Reflexive anaphora} & & & \\
        
        Person \& number agreement
        & \underline{0.4876} & \underline{\textbf{0.8716}} & 1737 \\

        Case agreement
        & 0.8534 & \textbf{0.9691} & 648 \\

        \bottomrule
    \end{tabular}
    \caption{Performances (accuracy) of two TLMs on German syntactic agreement tests. Underlined are the five tasks the models performed worst on. Bold-faced are the best scores per task.}
    \label{tab:experiments}

\end{table*}

Another agreement test relates to the more diverse word order in German in comparison to English. 
Example~\ref{exmp:vorfeld} illustrates the shift of the direct object \textit{diese Romane} from its standard position in the middle-field (after the finite verb) to the pre-field, and the shift of the subject \textit{der Autor} to the middle-field from its standard position in the pre-field (before the finite verb). This movement would be not possible in English. The German language often allows the shift, since it marks the case of noun phrases by the inflectional suffix of their determiner (e.g. \textit{der Autor} in nominative case vs. \textit{den Autor} in accusative case) and sometimes also by the suffix of the noun itself (e.g. \textit{des Autors} in genitive). That property supports to distinguish subjects (always nominative case) from objects or adjuncts independent from their position in a sentence. With this test case, we can evaluate if the model recognizes the subject in sentences correctly, even though the subject-verb-object order is disregarded.
We exclude test sentences where the subject and the object have the same inflectional suffixes in nominative and accusative, i.e. an unambiguous distinction between subject and object is not possible solely based on the inflection.

\begin{exmp}
\label{exmp:vorfeld}
\textit{Pre-field}
\begin{enumerate}[topsep=0px,noitemsep,label=(\alph*)] 
    \item Diese Romane \textit{empfahl} der Autor.
    \item * Diese Romane \textit{empfahlen} der Autor. 
\end{enumerate}
\end{exmp}

Moreover, we created sentences with reflexive verbs, i.e. sentential phrases where the reflexive anaphora (RA) in the accusative case follows the verb and agrees with the subject in the grammatical number and person. The first sentence in Examples~\ref{exmp:reflpers} and~\ref{exmp:reflcase} illustrates the agreement between RA \textit{mich} (accusative case) and the subject \textit{ich} in person (first) and number (singular).  We use two different tests: (a) for the recognition of a correct person (Example~\ref{exmp:reflpers}), also used by \citet{marvin-linzen-2018}, and (b) for the recognition of a correct case (accusative instead of incorrect dative, Example~\ref{exmp:reflcase}). The correct number is always given.

\begin{exmp}
\label{exmp:reflpers}
\textit{Subject RA agreement (person-agreement)}
\begin{enumerate}[topsep=0px,noitemsep,label=(\alph*)] 
        \item  Ich bedanke \textit{mich}.
        \item * Ich bedanke \textit{sich}.
\end{enumerate}
\end{exmp}

\begin{exmp}
\label{exmp:reflcase}
\textit{RA in accusative (case-agreement)}
\begin{enumerate}[topsep=0px,noitemsep,label=(\alph*)] 
        \item Ich bedanke \textit{mich}. 
        \item * Ich bedanke \textit{mir}.
\end{enumerate}
\end{exmp}

%% file: sections/4_experiments.tex
\section{Experiments}

In this section, we introduce the models we evaluate and in particular highlight their similarities and differences. We probe transformer-based \textsc{Bert} models because they are currently the basis for many state-of-the-art downstream models and very prominent in the community. The model selection was driven and confined by availability. We made use of \citet{Wolf2019HuggingFacesTS}'s transformers package.\footnote{\url{https://github.com/huggingface/transformers} (Accessed: 2020-03-05)}

The first model which we refer to as \textsc{gBert}$_{\text{large}}$ is a community model provided by the Bavarian State Library.\footnote{\url{https://huggingface.co/dbmdz/bert-base-german-cased} (Accessed: 2020-03-05)} It was trained on multiple German corpora including a recent Wikipedia dump, EU Bookshop corpus, the Open Subtitles corpus, a CommonCrawl corpus, a ParaCrawl corpus and the News Crawl corpus, with 16 GB of training material in total.
\sloppy
The second model which we refer to as $_{\text{distil}}$\textsc{gBert} was trained on half of the data used to pretrain \textsc{Bert} using distillation with the supervision of \textsc{gBert}$_{\text{large}}$ \footnote{\url{https://github.com/huggingface/transformers/blob/master/examples/distillation/README.md} (Accessed: 2020-05-21)}.

The data set, the CFGs with the list of lexical items and the code for the experiments are publicly available.~\footnote{\url{https://github.com/DFKI-NLP/gevalm/}}

%% file: sections/5_resultsanddiscussion.tex
\section{Results \& Discussion}

\begin{table*}[htbp]
    \centering
    \scalebox{0.85}{
    \begin{tabular}{rccc}
        \toprule
        & $_{\text{distil}}$\textsc{gBert} & \textsc{gBert}$_{\text{large}}$ & \# sents \\

        \textsc{Subject-verb agreement} & & & \\
        \midrule

        \multirow{2}{*}{Simple sentence}
        -sg & 0.9744  & 0.8974 & 39 \\
        -pl & 0.9667 & 1.0 & 30 \\
        
        \midrule
        
        \multirow{4}{*}{In a sentential complement}
        -sgsg & 1.0 & 0.9593 & 540 \\
        -plpl & 0.8926  & 1.0 & 270 \\
        -sgpl  & 0.9407 & 1.0 & 1080 \\
        -plsg & 0.9963 & 0.9963 & 270\\
        
        \midrule
        
        \multirow{2}{*}{Short VP coordination} 
        -sg & 0.8917 & 0.9667 & 120 \\
        -pl & 0.5333  & 0.5417 & 120 \\
        
        \midrule
        
        \multirow{4}{*}{Medium VP coordination} 
        -sgsg & 0.7667 & 0.95 & 120 \\
        -plpl & 0.2333  & 0.3167 & 120 \\
        -sgpl & 0.775 & 0.9667 & 120 \\
        -plsg & 0.15 & 0.2417 & 120 \\
        
        \midrule
        
        \multirow{4}{*}{Long VP coordination}
        -sgsg & 0.5917  & 1.0 & 120 \\
        -plpl & 0.2  & 0.2167 & 120 \\
        -sgpl & 0.4917  & 1.0 & 120 \\
        -plsg & 0.7833  & 0.1583 & 120 \\
        
        \midrule

        \multirow{4}{*}{Across a prepositional phrase}
        -sgsg & 0.7593  & 0.8667 & 540 \\
        -plpl & 0.7759  & 0.9426 & 540 \\
        -sgpl & 0.7907 & 0.8333 & 540 \\
        -plsg & 0.8611 & 0.9593 & 540 \\
        
        \midrule

        \multirow{4}{*}{Across a subject relative clause}
        -sgsg & 0.4222  & 0.9944 & 360 \\
        -plpl & 1.0  & 0.975 & 360 \\
        -sgpl & 0.3638  & 0.9889 & 360 \\
        -plsg & 0.9833  & 1.0 & 360 \\
        
        \midrule

        \multirow{4}{*}{Across an object relative clause}
        -sgsg & 0.4148 & 0.963 & 270 \\
        -plpl & 0.9667 & 0.9481 & 270 \\
        -sgpl & 0.4889 & 0.7481 & 135 \\
        -plsg & 0.9593 & 0.937 & 270 \\
        
        \midrule

        \multirow{4}{*}{In an object relative clause}
        -sgsg & 0.9911  & 1.0 & 450 \\
        -plpl & 0.9511  & 0.9422 & 450\\
        -sgpl  & 0.88 & 0.9822 & 225 \\
        -plsg & 0.9667 & 0.9267 & 450\\
        
        \midrule

        \multirow{2}{*}{With a simple modifier}
        -sg & 0.975  & 1.0 & 120 \\
        -pl & 0.9167 & 0.9917 & 120 \\
        
        \midrule

        \multirow{4}{*}{With an extended modifier}
        -sgsg & 0.9417 & 0.9667 & 120 \\
        -plpl & 0.8 & 0.9583 & 120 \\
        -sgpl & 0.9083 & 0.9667 & 120 \\
        -plsg & 0.9167 & 0.9417 & 120 \\
        
        \midrule

        \multirow{4}{*}{Pre-field}
        -sgsg & 0.7167 & 0.975 & 120\\
        -sgpl & 0.574 & 0.6759 & 120 \\
        -plsg & 0.8833 & 0.7333 & 108 \\
        
        \toprule

        \textsc{Reflexive anaphora} & & & \\
        \midrule

        \multirow{3}{*}{Person \& number agreement}
        -simple & 0.3611 & 0.6389 & 72 \\
        -longer & 0.3492 & 0.7841 & 315 \\
        -SentCompl & 0.5267 & 0.9045 & 1350\\
        
        \midrule
        
        \multirow{3}{*}{Case agreement}
        -simple & 0.9444 & 1.0 & 18 \\
        -longer & 0.7222 & 0.7889 & 90\\
        -SentCompl & 0.8722 & 0.9981 & 540 \\

        \bottomrule
    \end{tabular}}
    \caption{Fine-grained results of our experiments. Double-case specifications, e.g. "-plsg", are to be read as the tested agreement being in plural form, while the distractor is in singular form.}
    \label{tab:finegrained}
\end{table*}

The coarse-grained results of our experiments are listed in Table~\ref{tab:experiments}. We note that both models perform well across the majority of tasks. This is in line with previous work that demonstrated that \textsc{Bert} models are capable of solving syntactic agreement tasks. As shown by \citet{goldberg_assessing_2019} for English, for instance, our most successful German \textsc{Bert} model, \textsc{gBert}$_{\text{large}}$, also scores above 80\% or 90\% in most of the tasks, whereas the LSTM-LMs probed by \citet{marvin-linzen-2018} achieve scores not above 74\%. 
\sloppy
We observe that \textsc{gBert}$_{\text{large}}$ outperforms $_{\text{distil}}$\textsc{gBert} in thirteen out of fourteen tasks. For example, in the case of \textit{SVA across an object relative clause}, \textsc{gBert}$_{\text{large}}$ achieved a score of 92.06\%, whereas $_{\text{distil}}$\textsc{gBert}'s score is lower by around 18 percentage points. Based on these observations, we assume the higher amount of German training data, that \textsc{gBert}$_{\text{large}}$ was trained on, is the distinguishing factor.

There is a big overlap between the most challenging stress tests. Four out of five tests align when sorted in ascending order (worst performance first, underscored in Table~\ref{tab:experiments}). To analyse the stress tests further, in Table~\ref{tab:finegrained}, we subdivide cases between singular and plural subjects and distractors.

We expected high accuracies for the cases with local agreement. Our results show that all those cases, which are \textit{Simple Sentence}, \textit{SVA in a sentential complement}, \textit{SVA in an object relative clause} and \textit{SVA with a simple modifier}, have a score above 94 percent for both models.

Regarding the German-specific syntactic constructs, we observe that both models perform well. 
The movement of the subject from pre-field to middle-field does not seem to cause any bigger problems, both $_{\text{distil}}$\textsc{gBert} and \textsc{gBert}$_{\text{large}}$ have an accuracy between 0.73 and 0.8.

As can be seen in Tables~\ref{tab:experiments} and \ref{tab:finegrained}, VP coordination probing cases were a big challenge for both models. For example, $_{\text{distil}}$\textsc{gBert} only achieves an overall accuracy of 0.4813 on \textit{SVA in a medium VP coordination} and 0.5167 on \textit{SVA in a long VP coordination}, while \textsc{gBert}$_{\text{large}}$ achieves 0.6188 and 0.5938, respectively. In these aspects, our results deviate considerably from the findings of \citet{goldberg_assessing_2019} who reported that the English \textsc{Bert} models performed well on long VP tasks, too. The respective syntactic constructs may thus be particularly challenging for the \textsc{Bert} models in the German language. Interestingly, according to Table \ref{tab:finegrained}, \textsc{gBert}$_{\text{large}}$ performs with an accuracy of 1.0 for long VPs with a singular subject. We note that the most challenging sentences for both models in all of the VP coordination cases were the ones with a plural subject.

In contrast to the aforementioned VP coordinations, \textit{SVA across an object relative clause} for both models and  \textit{SVA across a subject relative clause} for $_{\text{distil}}$\textsc{gBert} show a better accuracy for sentences when the subject is plural. We assume that for some cases the grammatical number of the subject is a more influential aspect for the result than the number of the distractor. We didn't expect this given that we used the same lexemes within one case to ensure comparability between the results.

We expected that sentences in which the grammatical number of the distractor deviates from the number of the relevant verb (singular-plural and plural-singular) have a lower accuracy.
This, however, applies only to a few cases, like \textit{SVA across an object relative clause} and \textit{Pre-field}.
Thus, the TLMs appear to be mostly robust against those distractors.

Inferring sound causes for why some syntactic constructs push the models to their limit would require a thorough statistical analysis of the data and probably even an introspective analysis of the model. We leave it to future work to conduct such an analysis.

%% file: sections/6_related_work.tex
\section{Related Work}
There is a huge body of related literature on the syntactic evaluation of language models. For more background, we refer the interested reader to the works cited in the influential contribution by \citet{marvin-linzen-2018} and \citet{goldberg_assessing_2019}.

\citet{gulordava_colorless_2018} assessed subject-verb agreement with an emphasis on syntactic over semantic preference. \citet{mccoy_right_2019} created a data set with entailment tests. \citet{bacon_does_2019} extended \citet{goldberg_assessing_2019} to 26 languages, excluding German, and found out that with a higher number of distractors and long-range dependencies, \textsc{Bert} achieves lower accuracies for the syntactic agreement tests.

As mentioned above, we also recommend the overview paper by \citet{rogers2020primer} on studies of \textsc{Bert} models specifically. Apart from  the experiments cited in this work that evaluate multi-lingual models, such as \textsc{mBert}, we are not aware of any study dedicated to the agreement analysis of German \textsc{Bert} models. 

\citet{ronnqvist_is_2019}, nevertheless, tested multilingual \textsc{Bert} models on their ability of hierarchical understanding of German sentences and with a cloze test for which an arbitrary (grammatically correct) word was masked and needed to be filled in again.

%% file: sections/7_conclusion.tex
\section{Conclusion}

We conducted a broad analysis of German \textsc{Bert} models, targeting their syntactic abilities. We translated agreement tests from English to German and also designed tasks that reflect syntactic phenomena that are typical for the German language. The data set we generated and the accompanying grammars are publicly available. 

Furthermore, we utilized the cross-entropy loss to score whole natural sentences and this way mitigated a problem with sub-word tokenization. Our source code is open source, too.

Our experimental results show that the German models perform very well on most of the agreement tasks. However, we also identified syntactic stress tests that models in other languages appear to solve much better. We plan to replace the synthetic sentences with real language samples in the future, to better reflect the diversity of the German language in our experiments. 

%% file: sections/acknowledgements.tex
\section*{Acknowledgements}
We would like to thank Leonhard Hennig for his valuable feedback. This work has been supported by the German Federal Ministry of Education and Research as part of the project \textsc{BBDC\_II (no.~01IS18025A)}.

%% file: sections/references.tex
\bibliographystyle{acl_natbib}
\bibliography{bibliography}